\documentclass[conference,compsoc]{IEEEtran}

\ifCLASSOPTIONcompsoc
  \usepackage[nocompress]{cite}
\else
  \usepackage{cite}
\fi

\ifCLASSINFOpdf
\else
\fi

\usepackage{graphicx}
\usepackage{amsmath}
\usepackage{amssymb}
\usepackage{booktabs}
\usepackage{epsfig}
\usepackage{enumitem}
\setlist[itemize]{leftmargin=3mm}
\usepackage{multirow}
\usepackage{makecell}
\usepackage{xr}
\usepackage{tikz}
\usepackage{hyperref}
\usepackage{dsfont}
\usepackage[ruled,vlined,linesnumbered]{algorithm2e}
\usepackage{algorithm2e}
\usepackage{algorithmicx}  
\usepackage{algpseudocode}



\def\eg{\emph{e.g.,}\xspace}
\def\etc{\emph{etc}\xspace}
\def\ie{\emph{i.e.,}\xspace}

\DeclareMathOperator{\SAM}{\texttt{SAM}}
\DeclareMathOperator{\TAM}{\texttt{TAM}}
\DeclareMathOperator{\styletransfer}{\texttt{ST}}
\DeclareMathOperator{\finetune}{\texttt{Finetune}}

\begin{document}

\title{LocalStyleFool: Regional Video Style Transfer Attack Using Segment Anything Model}

\IEEEoverridecommandlockouts
\author{
	\IEEEauthorblockN{
		Yuxin Cao\IEEEauthorrefmark{2}, 
		Jinghao Li\IEEEauthorrefmark{3}, 
		Xi Xiao\IEEEauthorrefmark{2}\thanks{* Corresponding authors.}\IEEEauthorrefmark{1}, 
		Derui Wang\IEEEauthorrefmark{4}, 
            Minhui Xue\IEEEauthorrefmark{4}, 
            Hao Ge\IEEEauthorrefmark{5}, 
            Wei Liu\IEEEauthorrefmark{6}, 
            Guangwu Hu\IEEEauthorrefmark{6}\IEEEauthorrefmark{1}} 
	\IEEEauthorblockA{\IEEEauthorrefmark{2}Shenzhen International Graduate School, Tsinghua University, China}
	\IEEEauthorblockA{\IEEEauthorrefmark{3}Shandong University, China}
	\IEEEauthorblockA{\IEEEauthorrefmark{4}CSIRO's Data61, Australia} 
	\IEEEauthorblockA{\IEEEauthorrefmark{5}Ping An Technology, China}
        \IEEEauthorblockA{\IEEEauthorrefmark{6}Shenzhen Institute of Information Technology, China}
}

\maketitle

\begin{abstract}
Previous work has shown that well-crafted adversarial perturbations can threaten the security of video recognition systems. Attackers can invade such models with a low query budget when the perturbations are semantic-invariant, such as StyleFool. Despite the query efficiency, the naturalness of the minutia areas still requires amelioration, since StyleFool leverages style transfer to all pixels in each frame. To close the gap, we propose LocalStyleFool, an improved black-box video adversarial attack that superimposes regional style-transfer-based perturbations on videos. Benefiting from the popularity and scalably usability of Segment Anything Model (SAM), we first extract different regions according to semantic information and then track them through the video stream to maintain the temporal consistency. Then, we add style-transfer-based perturbations to several regions selected based on the associative criterion of transfer-based gradient information and regional area. Perturbation fine adjustment is followed to make stylized videos adversarial. We demonstrate that LocalStyleFool can improve both intra-frame and inter-frame naturalness through a human-assessed survey, while maintaining competitive fooling rate and query efficiency. Successful experiments on the high-resolution dataset also showcase that scrupulous segmentation of SAM helps to improve the scalability of adversarial attacks under high-resolution data.
\end{abstract}

\IEEEpeerreviewmaketitle

\section{Introduction}\label{sec:intro}
With the tentacles of deep learning extending to various computer vision applications in daily life, \eg autonomous driving~\cite{yurtsever2020survey}, smart home~\cite{bianchi2019iot} and virtual/augmented reality~\cite{zhu2020haptic}, its success and convenience have benefited most ordinary people. However, everything is a double-edged sword: Deep Neural Networks (DNNs) are found to be vulnerable to adversarial examples generated by minuscule imperceptible perturbations that can mislead the classifier~\cite{szegedy2014intriguing,goodfellow2015explaining}. This decade-long discovery has posed a threat to the security of DNNs in both academia and industry. Despite the popularity of short videos in today's era, video recognition models cannot escape by sheer luck from adversarial attacks. Therefore, comprehensive research on video adversarial attacks is in great demand.

In the early stage, studies are centered on restricted adversarial attacks in which perturbations are constrained within the $\ell_p$ norm ball around the clean input~\cite{wei2019sparse,jiang2019black,li2019stealthy,li2021geometric}. Such attacks keep the requirement of image attacks that perturbations are imperceptible both mathematically and visually. However, the negligible perturbations require large amounts of queries when the adversary has no access to the surrogate model (black-box setting). Recently, some scholars have proposed a new branch of attack: unrestricted attacks, where the adversary can enlarge the perturbation size to make the perturbations as imperceptible as possible. This kind of semantic-invariant attack can be roughly divided into patch-based attacks and style-based attacks. Despite the efficiency improvement and strong capability to bypass adversarial defenses, there are significant flaws in naturalness and temporal consistency. Patches such as bullet-screen comments~\cite{chen2022attacking}, and semantically unrelated stickers~\cite{jiang2023STDE}, are obvious enough to alert people. StyleFool~\cite{cao2023stylefool} is the first attempt to introduce style transfer to video attack. It elaborately designs a style selection strategy to change the style of all pixels in each frame, ensuring temporal consistency and overall naturalness. However, such practice inevitably brings about a local color abnormality that is counterfactual to human cognition, \ie green skin. 

\begin{figure*}[t]
\begin{center}
  \includegraphics[width=0.8\linewidth]{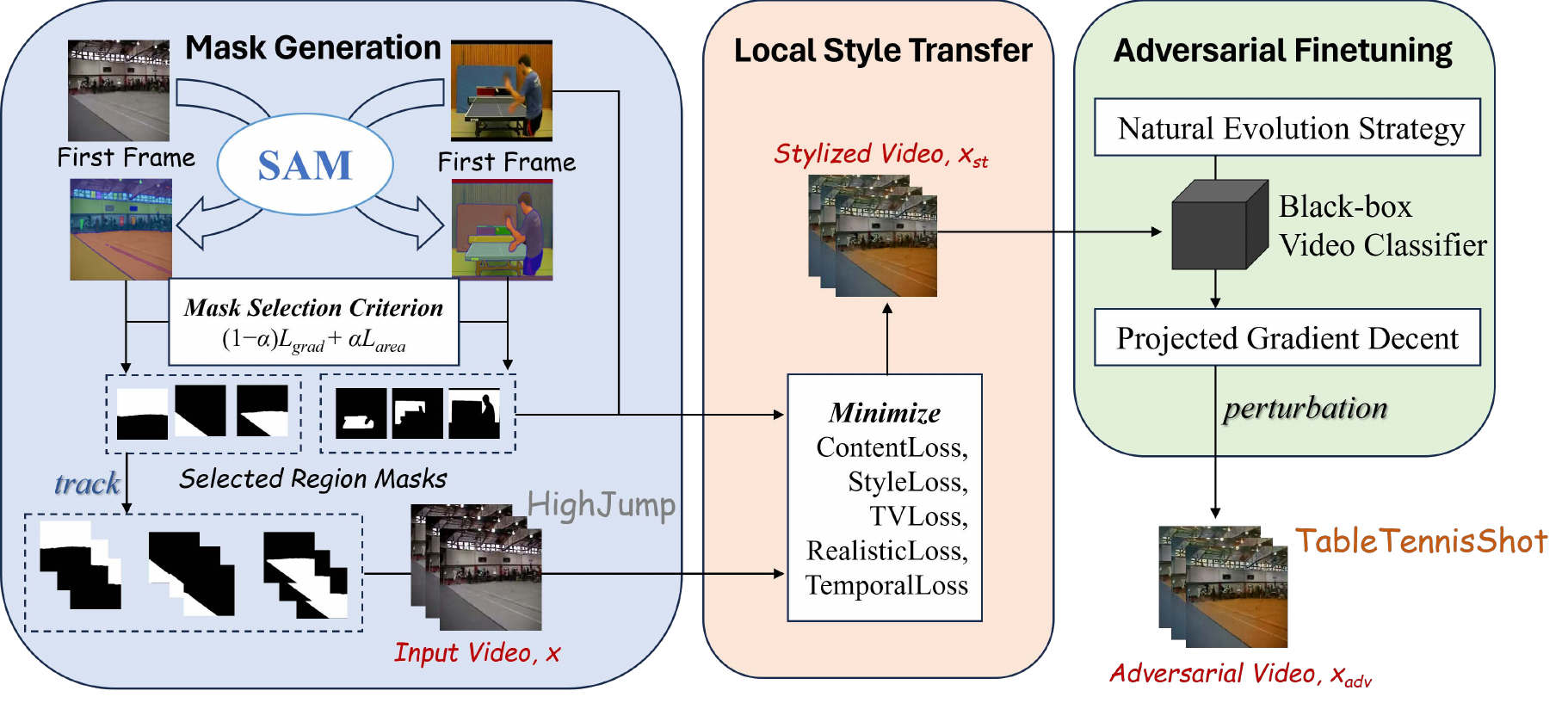}
\end{center}
\vspace{-2mm}
  \caption{Overview of LocalStyleFool.}
\label{fig:overview}
\vspace{-2mm}
\end{figure*}

To bridge the gap mentioned above, we propose a novel black-box attack for video recognition models called LocalStyleFool. As an improvement of StyleFool, our attack considers adding style transfer in several regional areas using different but natural style images. Concretely, we first use the Segment Anything Model (SAM)~\cite{kirillov2023SAM} to extract different semantic regions, and then find several important regions which are at the top of an associative criterion of the Grad-CAM (Gradient-weighted Class Activation Mapping)~\cite{selvaraju2017gradcam,gradcam_pytorch} output for a local pre-trained image model and the regional area. Afterwards, these regions are tracked through the video stream and transferred by different style images extracted from the target class video. Finally, similar to StyleFool, the perturbations are fine-adjusted to ensure misclassification. We conduct a user study to showcase the advantage of both intra-frame and inter-frame naturalness compared with StyleFool, while maintaining competitive attack efficiency. Experiments on a high-resolution video dataset, Kinetics-700~\cite{carreira2019kinetics700}, also show that refined style-transfer-based perturbations in high-resolution videos can help improve video quality and avoid local counterfactual details. Despite the positive benefits of SAM, we manifest that it can also be used to malicious or illegal behavior, such as forging adversarial examples.

Our main contributions are summarized as follows.
\begin{itemize}
    \item We improve StyleFool and propose LocalStyleFool, a new black-box video adversarial attack on video models, leveraging style transfer to different semantic regions using different style images. This closes the gap in the naturalness insufficiency of StyleFool.
    \item LocalStyleFool requires competitive query budgets but improves regional naturalness in the frame and temporal consistency across frames, which is verified by comprehensive experiments and a human-accessed survey.
    \item LocalStyleFool also witnesses excellent attack performance on the high-resolution video dataset, catering to the video development trend. To the best of our knowledge, we are the first to employ the Segment Anything Model to conduct malicious adversarial attacks. We hope to increase the attention of the security community to the negative impacts of such techniques.
\end{itemize}

\subsection{Restricted Video Attacks}
Early study focused on the white-box setting, where the structure and parameters of the model are available to the adversary~\cite{wei2019sparse,li2019stealthy,wu2023imperceptible}. The perturbations were required to ensure stealthiness under the $\ell_p$ norm restriction. Since the black-box setting has won better popularity and is more pragmatic, numerous black-box attacks have come into sight in academia these years~\cite{jiang2019black,wei2020heuristic,li2021geometric,wei2022adaptive}. In addition, reinforcement learning has been widely used to optimize high-dimensional perturbations or extract key frames and regions to enhance sparsity~\cite{wang2021reinforcement,yan2021efficient,wei2022sparse,wei2023efficient}. There is also some work on cross-modal attacks that generated video perturbations from images/image models~\cite{kim2023breaking,wang2023global}. 
However, redundant queries lead to low attack efficiency.

\subsection{Unrestricted Video Attacks}
To make the perturbations larger and reduce the query budget, some research considered unrestricted perturbations which are not related to video semantics. This branch of attacks includes changing style~\cite{cao2023stylefool,cao2024logostylefool} and adding patches~\cite{jiang2023STDE}, flickers~\cite{pony2021flickering}, extra frames~\cite{chen2021appending} or bullet-screen comments~\cite{chen2022attacking}. StyleFool~\cite{cao2023stylefool} fools video classifiers by introducing natural style transfer to initialize the input video. However, the overall style transfer brings about color and texture distortion in local areas, which is not in line with human cognition and natural laws. In order to make perturbations more natural, Chen et al.~\cite{chen2022attacking} added large bullet-screen comments to the video, but this attack is not suitable for targeted attacks, which was later verified by Jiang et al.~\cite{jiang2023STDE} They added patch-based perturbations to the video, but the patches are not semantically related to the original input video, and the perturbation range is so large that human perception systems can easily be aware of the perturbations. Also, some large patches directly obscure the core action area of the video. Overall, most unrestricted perturbations were much too obvious in the level of both intra-frame (\eg weird patches, counterfactual details) and inter-frame (\eg erratic transitional flicker), bringing about lack of naturalness and temporal inconsistency. 

\section{Method}
\subsection{Preliminary}
We assume a DNN video recognition model $f$ which takes a video $x$ with $T$ frames as input. The output of the model $f$ includes the predicted label $y$ and its confidence score. We assume that the adversary launches the attack under a query-limited black-box setting where only the top-1 label and its score are accessible, and there is a query limit for the attack, since the higher query budget lacks practicality in real-world scenarios. Moreover, the adversary can choose the attack goal from targeted attacks and untargeted attacks. The adversarial video $x_{adv}$ should be misclassified to a predefined target class $y_t$ in targeted attacks, \ie $f\left( x_{adv} \right) = y_t$, while to any class other than the ground-truth class $y_0$ in untargeted attacks, \ie $f\left( x_{adv} \right) \ne y_0$. We provide the overview of LocalStyleFool in Figure~\ref{fig:overview}.

\subsection{Mask Generation}
Adversarial videos generated by StyleFool (as shown in Figure~\ref{fig:visualization_compare}) are found unnatural in regional details, since StyleFool applies style transfer to all pixels of each frame. To improve this gap, an intuitive alternative is to introduce different style transfer to different regions. However, a pilot study conducted ourselves shows that it is rather difficult to segment different semantic regions in the resized videos. This conclusion is based on two fundamental facts. 1) Existing video attacks design and optimize perturbations after resizing the video to the input size requirement of the surrogate model. It is of nature and reality, since the video that is projected to misclassify the model is the resized video. 2) Most existing widely used video recognition models (\eg C3D~\cite{ji2012c3d}, I3D~\cite{carreira2017i3d}) require the input videos to be resized to a low resolution (lower than 224 $\times $ 224), making it difficult to apply object segmentation methods. These facts prompt us to consider designing perturbations on the original high-resolution videos. Thanks to the SAM~\cite{kirillov2023SAM} proposed recently, it is a relatively convenient alternative to introduce SAM to video adversarial attacks. 

Due to the advantages of zero-hot transferability and accurate segmentation in high-resolution images, SAM has been used in various fields~\cite{mazurowski2023segment}. In our task, we first use pre-trained SAM to extract the masks for the first frame of the video. According to some existing work focusing on adding sparse perturbations to videos~\cite{yan2021efficient,wei2022sparse}, different regions make different contributions to the prediction of the recognition model. Therefore, we choose some of the regions from the output masks of SAM to perform style transfer. To be specific, we design an associative criterion to choose the regions to be style-transferred.

\noindent\textbf{Transfer-based gradient information.} 
In black-box attacks, the adversary has no access to the model architecture. Due to the transferability of adversarial perturbations across models~\cite{papernot2016transferability} and data domains (from images to videos)~\cite{jiang2019black,kim2023breaking}, we utilize a local white-box image model to obtain the significance of different regions by Grad-CAM~\cite{selvaraju2017gradcam}. Concretely, we first use a pre-trained ResNet-50~\cite{he2016ResNet} model $g$ to obtain the heatmap for class $y_c$:
\begin{small}
\begin{equation}\label{equ:heatmap}
{H^{y_c}} = \mathrm{ReLU}\left( {\frac{1}{SR} \sum\limits_k {\sum_{i,j}\frac{\partial g \left(y_c|x_1 \right)}{\partial F_{i,j}^k}F^k\left( x_1 \right)} } \right),
\end{equation}
\end{small}%
where $\mathrm{ReLU}$ represents the ReLU activation function, $F^k$ represents the $k$-th feature map for the first frame $x_1$, ${\left( {i,j} \right)}$ denotes the pixel position of the image, $SR$ denotes the spatial resolution of the feature map. We use the first frame since one-action short video has no manifest difference across frames~\cite{cao2023stylefool}.

We use $M^p$ to represent the mask matrix for the $p$-th region of the output of SAM, which is a sparse matrix, in which the pixel belonging to this segmentation region is set to 1, otherwise 0. 
Assuming that SAM outputs $N$ regions ${{m_1},{m_2},...,{m_N}} $ (${m_i} \cap {m_j} = \emptyset,\forall i,j \in \left\{ {1,2,...,N} \right\},i \ne j$) and their corresponding masks ${{M^1},{M^2},...,{M^N}} $, the normalized gradient significance can be expressed as
\begin{small}
\begin{equation}
{L_{grad}}\left( p \right) = \frac{{\sum\limits_{\left( {i,j} \right) \in {m_p}} {M_{i,j}^pH_{i,j}^{{y_c}}}  - \mathop {\min }\limits_q \sum\limits_{\left( {i,j} \right) \in {m_q}} {M_{i,j}^qH_{i,j}^{{y_c}}} }}{{\mathop {\max }\limits_q \sum\limits_{\left( {i,j} \right) \in {m_q}} {M_{i,j}^qH_{i,j}^{{y_c}}}  - \mathop {\min }\limits_q \sum\limits_{\left( {i,j} \right) \in {m_q}} {M_{i,j}^qH_{i,j}^{{y_c}}} }},
\end{equation}
\end{small}%
where $m_p$ denotes the $p$-th region.

\noindent\textbf{Regional area.}
There is no direct relationship between the gradient significance and the regional area. In some cases, the areas of the top-ranked segmentation regions are very small, increasing the attack difficulty. That is to say, the cost-effectiveness of style transfer in these small regions is very low: complex calculations are required, but the impact on the classifier is minimal. To prioritize style transfer from regions with larger areas, we additionally consider the regional area as another criterion for choosing segmentation regions. Then, we consider the normalized area significance:
\begin{small}
\begin{equation}
{L_{area}}\left( p \right) = \frac{{{{\left\| {{M^p}} \right\|}_0} - \mathop {\min }\limits_q {{\left\| {{M^q}} \right\|}_0}}}{{\mathop {\max }\limits_q {{\left\| {{M^q}} \right\|}_0} - \mathop {\min }\limits_q {{\left\| {{M^q}} \right\|}_0}}}.
\end{equation}
\end{small}%

\noindent\textbf{Associative criterion.}
Since the gradient information and regional area contribute differently to the attack, we use the weighted significance criterion as follows.
\begin{small}
\begin{equation}\label{eq:criterion}
{L_{total}}\left( p \right) = \left( {1 - \alpha } \right){L_{grad}}\left( p \right) + \alpha {L_{area}}\left( p \right),
\end{equation}
\end{small}%
where $\alpha $ denotes the weight coefficient.
Then we obtain the masks $\mathcal{M}^1,\mathcal{M}^2,...,\mathcal{M}^N$ by sorting them in descending order according to $L_{total}$. However, the total area of the top-ranked regions is still different among videos. Therefore, to ensure that all adversarial videos have similar perturbed areas, we further set a total area lower bound. Concretely, we select the top-$r$ regions which satisfies that
\begin{small}
\begin{equation}
r = \mathop {\arg \max }\limits_{r'} r',s.t.\sum\limits_{p = 1}^{r' - 1} {{{\left\| {{{\cal M}^p}} \right\|}_0}}  \le \mu \sum\limits_{p = 1}^N {{{\left\| {{{\cal M}^p}} \right\|}_0}} 
\end{equation}
\end{small}%
where $\mu $ represents the total area lower bound coefficient. 

\subsection{Style Transfer}
Stylized videos will become unnatural and contrary to human cognition if the style images are selected randomly. Save for the visual naturalness brought by color, StyleFool uses target class confidence to move stylized videos close to the decision boundary, which shows that style images from the target class carry a lot of target class information. Therefore, we select the target class video with the highest confidence score in the targeted class, while a random video from the class other than the original class in untargeted attacks. To make stylized videos more natural, we consider the image with different style regions. We first use the associative criterion in Equation~\ref{eq:criterion} to obtain the top-$r$ masks ${\Psi ^1},{\Psi ^2},...,{\Psi ^r}$ for target class video $x_t$. Then, the $r$ regions of the original video $x$ will undergo style transfer using the style images $s$ as the $r$ regions of the target class video $x_t$. In content loss and style loss, we use VGG-19~\cite{simonyan2014very} to obtain high-level features. We improve the style loss to:

\begin{small}
\begin{equation}
{L_{{\rm{style }}}}\left( {{x^s},s} \right) = \sum\limits_l {\sum\limits_p {\frac{1}{{C_l^2}}\left\| {{G_{l,p}}\left( s \right) - {G_{l,p}}\left( {{x^s}} \right)} \right\|_2^2} },
\end{equation}
\end{small}%
where $s$ denotes the style image from the target class video, $x^s$ denotes the stylized video frame, $C_l$ denotes the channel number in the $l$-th layer, ${G_{l,p}}$ denotes the Gram matrix~\cite{justin2016perceptual} corresponding to the feature $Q_{l,p}$ of the VGG-19 model. For $s$ and $x^s$, the features are expressed as

\begin{small}
\begin{equation}
\begin{array}{l}
{Q_{l,p}}\left( s \right) = \Psi _l^p{Q_l}\left( s \right)\\
{Q_{l,p}}\left( {{x^s}} \right) = \mathcal{M}_l^p{Q_l}\left( {{x^s}} \right),
\end{array}
\end{equation}
\end{small}%
where the subscript $l$ represents the $l$-th layer. The masks are downsampled to match the feature map in the $l$-th layer.

We additionally consider total variance loss and realistic loss~\cite{luan2017deep} to improve the smoothness at the spatial level. The realistic loss is built based on Matting Laplacian~\cite{levin2007closed}, which can penalize unnatural distortion to produce stylized photorealistic video frames. Similar to StyleFool~\cite{cao2023stylefool}, we introduce temporal loss to maintain the temporal consistency and leverage the Natural Evolution Strategy (NES)~\cite{ilyas2018black} and Projected Gradient Decent (PGD)~\cite{madry2017towards} to finetune the perturbations after video style transfer. Note that the masks change through the video stream. We use the Track Anything Model (TAM)~\cite{yang2023track} to track the top-$r$ regions with masks ${{\mathcal{M}^1},{\mathcal{M}^2},...,{\mathcal{M}^r}}$ in the original video to maintain temporal consistency. The $r$ styles extracted from the target class video remain unchanged. Algorithm~\ref{alg:attack} briefly describes the procedure of LocalStyleFool.

\section{Experiments}
\subsection{Experimental Setup}
\noindent \textbf{Datasets.}
Similar to most work on video adversarial attack tasks~\cite{chen2022attacking,wu2023imperceptible,xu2022sparse,wei2022sparse,cao2023stylefool}, we select UCF-101~\cite{soomro2012ucf101} and HMDB-51~\cite{kuehne2011hmdb} as our datasets due to their popularity and comprehensiveness. UCF-101~\cite{soomro2012ucf101} consists of 13,320 videos from 101 classes. HMDB-51~\cite{kuehne2011hmdb} contains 6,849 videos from 51 classes. We also include Kinetics-700~\cite{carreira2019kinetics700}, which contains approximately 650,000 videos from 700 classes. The resolution of the videos in Kinetics-700 is higher than that of UCF-101 and HMDB-51, with more than 40\% videos having a resolution of 1280 $\times $ 720 or higher. 
All videos were compiled from YouTube and contained only one action.

\noindent \textbf{Models.}
We choose C3D~\cite{tran2015learning} and I3D~\cite{carreira2017i3d} as surrogate models for UCF-101 and HMDB-51 because they are used most frequently for video recognition tasks. C3D learns temporal features by 3D convolution, while I3D uses optical flow to obtain transitional information between consecutive frames. We train the models ourselves before the attack.
For Kinetics-700, we replace C3D with R3D~\cite{hara2018R3D} due to better recognition performance of 3D ResNet models. We use the pre-trained model provided for Kinetics-700~\cite{repo_kinetics700}. The test dataset accuracy for UCF-101 is 85.2\% (C3D) and 86.9\% (I3D), for HMDB-51 is 67.0\% (C3D) and 62.8\% (I3D), and for Kinetics-700 is 68.4\% (I3D) and 63.1\% (R3D).

\noindent \textbf{Metrics.}
Following StyleFool~\cite{cao2023stylefool}, we use Attack Success Rate (ASR) and Minimal Queries (minQ), Maximal Queries (maxQ), and Average Queries (AQ) as our main metrics to evaluate quantitative attack performance. ASR stands for the ratio of successful attacks where the video is misclassified. The metrics related to queries showcase the attack efficiency. Note that the attack performance is also largely influenced by imperceptibility, which can be determined by visual perception by human subjects.

\noindent
\begin{algorithm}[t]\small
\footnotesize
\caption{LocalStyleFool.}\label{alg:attack}
\KwIn{Black-box model $f$, input video ${x}$, target class ${y_t}$, significance criterion $L_{total}$, style transfer loss $L_{st}$.}
\KwOut{Adversarial video ${x_{adv}}$.}
$x_{\left( 1 \right)} \gets $ the first frame of $x$\;
$x_t \gets $ target class video\;
$s \gets $ the first frame of $x_t $\;
${{M^1},{M^2},...,{M^N}} \gets \SAM \left( x_{\left( 1 \right)} \right)$\;
${{\psi^1},{\psi^2},...,{\psi^N}} \gets \SAM \left( s \right)$\;
${{\mathcal{M}^1_{\left( 1 \right)}},{\mathcal{M}^2_{\left( 1 \right)}},...,{\mathcal{M}^r_{\left( 1 \right)}}} \gets $ top-$r$ masks sorted by $L_{total}$\;
$\mathcal{M}^{1,2,...,r}_{\left( 2 \right)}, ..., \mathcal{M}^{1,2,...,r}_{\left( T \right)} \gets \TAM \left( x, \mathcal{M}^{1,2,...,r}_{\left( 1 \right)}\right)$\;
$\mathcal{M} \gets \left\{ \mathcal{M}^{1,2,...,r}_{\left( 1 \right)}, \mathcal{M}^{1,2,...,r}_{\left( 2 \right)}, ..., \mathcal{M}^{1,2,...,r}_{\left( T \right)} \right\}$\;
${{\Psi^1},{\Psi^2},...,{\Psi^r}} \gets $ top-$r$ masks sorted by $L_{total}$\;
$\Psi \gets \left\{ {{\Psi^1},{\Psi^2},...,{\Psi^r}} \right\}$\;
$x_{st} \gets \styletransfer \left( \mathcal{M}, \Psi, x, s, L_{st} \right)$\;
$x_{adv} \gets \finetune \left( x_{st}, y_t, x_t, f \right)$.
\end{algorithm}
\vspace{-4mm}

\noindent \textbf{Competitors.}
Since there are not many unrestricted attacks for videos, we choose two which are close to ours, STDE~\cite{jiang2023STDE}, StyleFool~\cite{cao2023stylefool}. STDE superimposes part of the regions in other videos on the input video, which achieves fewer queries, but the naturalness of adversarial videos is largely sacrificed. The patch is abrupt, large in size, lacks relevance to the semantics of the original video, and may obscure the important content of the original video. Although StyleFool achieves high attack efficiency while maintaining overall imperceptibility, local counterfactual colors and unnatural details still exist if videos are watched carefully. Please refer to the appendix for more experimental details and discussion on comparison with other video attacks.

\noindent \textbf{Fairness.}
Due to different techniques and experimental setups in STDE and StyleFool, it is not scalable and meaningful to modify them to the same framework as LocalStyleFool, \eg aligning to the same $\ell_p$ norm. Enlarging the $\ell_p$ norm for STDE will lead to fewer queries but to a more obvious patch, while the $\ell_p$ norm for StyleFool varies according to different style images. As a result, we use the default parameters for STDE and StyleFool, and focus more on the imperceptibility of visualization.

\subsection{Results and Analyses}
\begin{figure*}[tbp]
    \centering
    \begin{minipage}{0.65\linewidth}
        \includegraphics[width=1.0\linewidth]{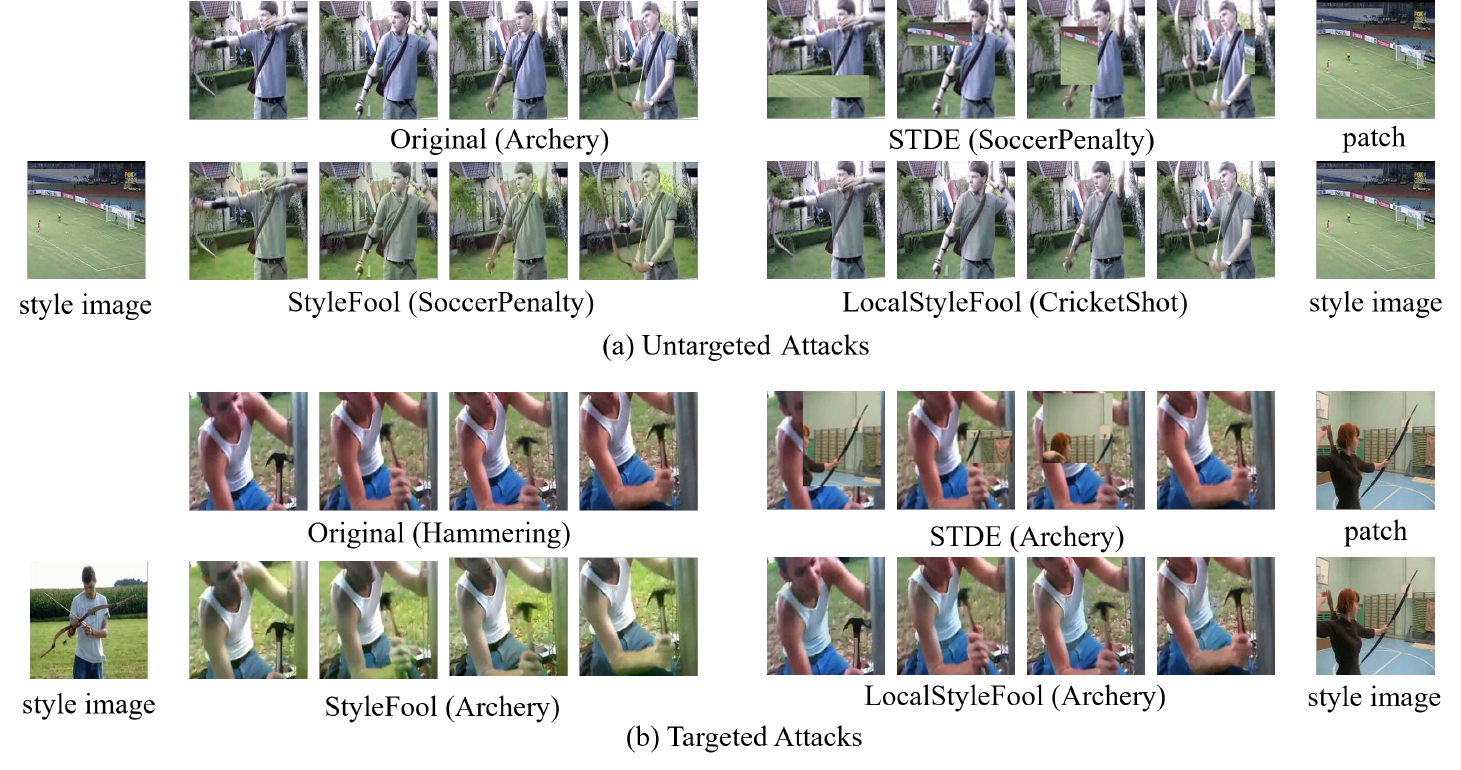}
        \vspace{-3mm}
        \caption{Visualization of different video attacks.}
        \label{fig:visualization_compare}
    \end{minipage}
    \quad
    \centering
    \begin{minipage}{0.3\linewidth}
        \includegraphics[width=1.0\linewidth]{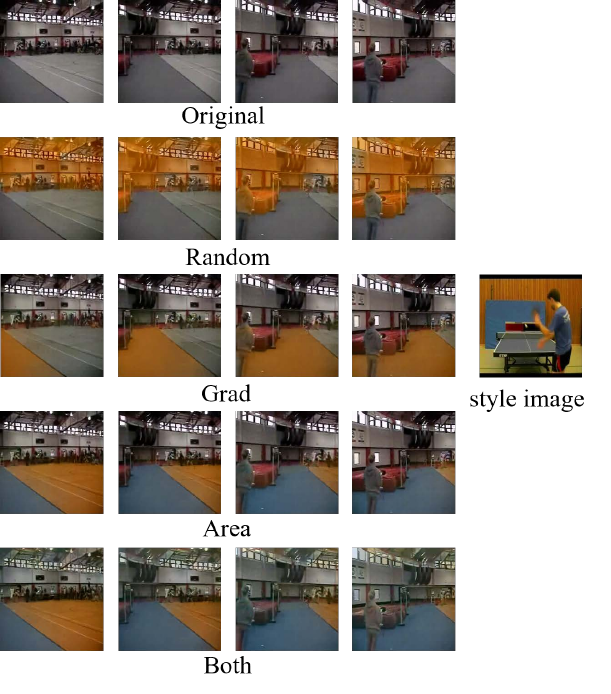}
        \vspace{-3mm}
        \caption{Visualization of ablation study.}
        \label{fig:visualization_ablation}
    \end{minipage}
\vspace{-2mm}
\end{figure*}

We randomly select 150 videos from all datasets and conduct LocalStyleFool as well as the other two competitors. Tables~\ref{tab:attack_performance_ucf101},~\ref{tab:attack_performance_hmdb51} and~\ref{tab:attack_performance_high_resolution} report the quantitative results. Note that the queries during the target class video selection are counted. Figure~\ref{fig:visualization_compare} shows the visualization. Since STDE adds large, unrestricted patches to the clean videos, the attack is the most query-efficient. However, the patches are so obtrusive that they even occlude the semantic region of the original action in the clean video. Even if STDE can quickly fool the video recognition model, this significant flaw in naturalness can directly lead to the failure of the attack: relevant personnel can easily perceive anomalies and trigger alarms. Compared to STDE, StyleFool requires more queries, but improves imperceptibility by a large margin. StyleFool transfers the clean video to another style for all pixels and, through an accurate and rigorous style selection strategy, achieves good naturalness. However, the overall style transfer can lead to unnatural regional details, \eg the green face and skin in Figure~\ref{fig:visualization_compare}. As an improvement, LocalStyleFool closes this gap and further enhances the naturalness by conducting style transfer on different regions, while maintaining comparable query cost. We provide examples of LocalStyleFool when conducting style transfer on one region in Figure~\ref{fig:visualization_compare} and multiple regions in Figure~\ref{fig:visualization_ablation} (last row). Results on Kinetics-700 indicate that performing style transfer on the high-resolution video first and then inputting it into the victim model is beneficial for improving attack efficiency. 

We also record the one-query attack success rate (1Q-ASR) to demonstrate the effectiveness of our method. The 1Q-ASR refers to the rate of clean videos that are immediately classified into target class (in targeted attacks) or any other class (in untargeted attacks) after style transfer. Thus, there is no need for the subsequent perturbation optimization. On average, LocalStyleFool achieves a 1Q-ASR of over 8\% in targeted attacks and 53\% in untargeted attacks. Especially, the 1Q-ASR achieves 81.3\% when attacking R3D on Kinetics-700 in the untargeted setting, which means that video recognition systems are extremely vulnerable to non-semantic perturbations which can push the stylized videos across the decision boundary at ease. We provide the discussion on potential mitigation in the appendix.

\begin{table}[t]  
\centering
\caption{{Attack performance comparison on UCF-101.}}
\label{tab:attack_performance_ucf101}
\resizebox{0.9\linewidth}{!}{
\begin{tabular}{ccrrrrrrrrrrrrrrrr}
\toprule
\multirow{2}{*}[-0.5ex]{\textbf{Model}} & 
\multirow{2}{*}[-0.5ex]{\textbf{Attack}} & \multicolumn{4}{c}{\textbf{UCF-101 (Targeted)}} & \multicolumn{4}{c}{\textbf{UCF-101 (Untargeted)}} \\
\cmidrule(r){3-6}\cmidrule(r){7-10}
& & \footnotesize{\textbf{ASR}} & \footnotesize{\textbf{minQ}} & \footnotesize{\textbf{maxQ}} & \footnotesize{\textbf{AQ}} &  \footnotesize{\textbf{ASR}} & \footnotesize{\textbf{minQ}} & \footnotesize{\textbf{maxQ}} & \footnotesize{\textbf{AQ}} \\
\midrule
\multirow{3}{*}{C3D} 
& STDE~\cite{jiang2023STDE} & 100 & 91 & 5,684 & 2,910 & 100 & 16 & 4,320 & 2,820 \\
& StyleFool~\cite{cao2023stylefool} & 100 & 1,322 & 273,446 & 73,104 & 100 & 1 & 18,772 & 3,676 \\
& LocalStyleFool & 100 & 1,274 & 215,932 & 70,425 & 100 & 1 & 19,072 & 3,575 \\

\midrule
\multirow{3}{*}{I3D} 
& STDE~\cite{jiang2023STDE} & 100 & 30 & 4,638 & 1,653 & 100 & 30 & 3,688 & 1,532 \\
& StyleFool~\cite{cao2023stylefool} & 100 & 101 & 122,740 & 32,074 & 100 & 1 & 29,517 & 6,557 \\
& LocalStyleFool & 100 & 101 & 117,481 & 33,472 & 100 & 1 & 31,409 & 7,052 \\
\bottomrule
\end{tabular}
}
\vspace{-2mm}
\end{table}

\begin{table}[t]  
\centering
\caption{Attack performance comparison on HMDB-51.}
\label{tab:attack_performance_hmdb51}
\resizebox{0.9\linewidth}{!}{
\begin{tabular}{ccrrrrrrrrrrrrrrrr}
\toprule
\multirow{2}{*}[-0.5ex]{\textbf{Model}} & 
\multirow{2}{*}[-0.5ex]{\textbf{Attack}} &\multicolumn{4}{c}{\textbf{HMDB-51 (Targeted)}} & \multicolumn{4}{c}{\textbf{HMDB-51 (Untargeted)}}\\
\cmidrule(r){3-6}\cmidrule(r){7-10}
& & \footnotesize{\textbf{ASR}} & \footnotesize{\textbf{minQ}} & \footnotesize{\textbf{maxQ}} & \footnotesize{\textbf{AQ}} &  \footnotesize{\textbf{ASR}} & \footnotesize{\textbf{minQ}} & \footnotesize{\textbf{maxQ}} & \footnotesize{\textbf{AQ}}  \\
\midrule
\multirow{3}{*}{C3D} 
& STDE~\cite{jiang2023STDE} & 100 & 30 & 5,598 & 2,165 & 100 & 16 & 3,754 & 1,849 \\
& StyleFool~\cite{cao2023stylefool} & 100 & 101 & 98,715 & 38,804 & 100 & 1 & 10,998 & 2,032 \\
& LocalStyleFool & 100 & 101 & 104,355 & 41,833 & 100 & 1 & 18,474 & 2,438 \\

\midrule
\multirow{3}{*}{I3D} 
& STDE~\cite{jiang2023STDE} & 100 & 30 & 4,896 & 1,623 & 100 & 19 & 6,879 & 1,835 \\
& StyleFool~\cite{cao2023stylefool} & 100 & 101 & 79,418 & 24,078 & 100 & 1 & 6,510 & 2,290 \\
& LocalStyleFool & 100 & 101 & 95,330 & 23,174 & 100 & 1 & 8,372 & 1,614 \\
\bottomrule
\end{tabular}
}
\vspace{-2mm}
\end{table}

\begin{table}[t]  
\centering
\caption{Attack performance comparison on Kinetics-700.}
\label{tab:attack_performance_high_resolution}
\resizebox{0.9\linewidth}{!}{
\begin{tabular}{ccrrrrrrrrrrrrrrrr}
\toprule
\multirow{2}{*}[-0.5ex]{\textbf{Model}} & 
\multirow{2}{*}[-0.5ex]{\textbf{Attack}} & \multicolumn{4}{c}{\textbf{Kinetics-700 (Targeted)}} & \multicolumn{4}{c}{\textbf{Kinetics-700 (Untargeted)}} \\
\cmidrule(r){3-6}\cmidrule(r){7-10}
& & \footnotesize{\textbf{ASR}} & \footnotesize{\textbf{minQ}} & \footnotesize{\textbf{maxQ}} & \footnotesize{\textbf{AQ}} &  \footnotesize{\textbf{ASR}} & \footnotesize{\textbf{minQ}} & \footnotesize{\textbf{maxQ}} & \footnotesize{\textbf{AQ}} \\
\midrule
\multirow{3}{*}{R3D}
& STDE~\cite{jiang2023STDE} & 100 & 182 & 32,065 & 8,731 & 100 & 16 & 5,654 & 1,810 \\
& StyleFool~\cite{cao2023stylefool} & 100 & 101 & 78,455 & 19,204 & 100 & 1 & 8,538 & 724 \\
& LocalStyleFool &  100 & 101 & 80,335 & 17,098 & 100 & 1 & 7,265 & 598 \\
\midrule
\multirow{3}{*}{I3D}
& STDE~\cite{jiang2023STDE} & 100 & 107 & 48,962 & 15,651 & 100 & 30 & 6,547 & 3,205 \\
& StyleFool~\cite{cao2023stylefool} & 100 & 101 & 122,481 & 24,573 & 100 & 1 & 12,601 & 1,536 \\
& LocalStyleFool & 100 & 101 & 113,472 & 25,307 & 100 & 1 & 9,914 & 1,298 \\
\bottomrule
\end{tabular}
}
\vspace{-2mm}
\end{table}
 
\subsection{User Study}
Apart from quantitative analyses, we also conduct a human-centered census to show the indistinguishability of the proposed LocalStyleFool, which is a supplement to the visualization effect from a human perspective. We conducted an online survey on Amazon Mechanical Turk~\cite{amazon_turk} and recruited 100 anonymous subjects over 18 years of age. All subjects are from English-speaking countries and can complete the survey in English. Following most user studies in similar adversarial attack tasks~\cite{cao2023stylefool,diao2021basar}, our survey consists of three parts, naturalness, realness, and consistency. Subjects were paid \$0.8 for each question. 

\textbf{Naturalness:} The naturalness refers to the degree to which its color, texture, and overall appearance are consistent with human basic cognition. We randomly selected 20 clean videos and their corresponding adversarial counterparts for three attacks. The subjects were asked to evaluate the naturalness based on a Likert-scale~\cite{likert1932technique} from 1 to 5. A higher score indicates more natural. 

\textbf{Realness:} The realness of refers to the degree to which it appears to be shot realistically in the real world, with almost no traces of artificial processing. We then grouped the videos to ask the subjects to distinguish the realness. 

\textbf{Consistency:} The consistency includes spatial consistency (smoothness of value changes between a pixel and its surrounding pixels) and temporal consistency (smoothness and coherence of transitions between frames). We randomly selected another 40 videos, 10 from each attack, and asked the subjects to rate the consistency from 1 to 5. A higher score indicates greater consistency. The order of the videos was randomly shuffled to avoid potential bias.

\noindent \textbf{Analyses.} We finally obtained 97 valid questionnaires after filtering out 3 questionnaires whose complete time is much shorter than the total time for playing all videos. Table~\ref{tab:user_study} reports the average naturalness and consistency results obtained from subjects. Although STDE requires the least queries by adding large patches to the video, naturalness is greatly sacrificed, which can easily trigger alerts and finally lead to a futile attack. Patches are conspicuous and inserted into sparse frames, which also greatly reduces the consistency rate. StyleFool improves naturalness and consistency greatly since it considers fine style selection to ensure the stylized video not only maintains good imperceptibility but is also close to the decision boundary. However, style transferring on all pixels leads to counter-cognitive details in local areas such as aberrant colors and texture. LocalStyleFool closes this gap and obtains higher naturalness and consistency rates that are closest to those of clean videos.

Table~\ref{tab:realness} reports the realness test results. Compared to STDE, 51.5\% of the subjects thought LocalStyleFool is real while this figure was only 6.2\% for STDE. As analyzed previously, the unnatural patch of STDE reduces video quality and makes people feel that the videos have been maliciously modified. Over 40\% of the subjects consented that LocalStyleFool generated more real videos than StyleFool, showing the outstanding advantage of LocalStyleFool in maintaining local smooth details. Nearly 30\% of the subjects considered both videos were real, indicating that the style-transfer-based perturbations carry high stealthiness. It is surprising that up to 34.0\% of the subjects rooted for LocalStyleFool, while clean videos only obtained 37.1\% of the votes. This shows that LocalStyleFool can achieve high verisimilitude similar to clean videos.

\begin{table}[t]  
\centering
\caption{Results of naturalness and consistency tests.}
\label{tab:user_study}
\resizebox{0.6\linewidth}{!}{
\begin{tabular}{ccccc}
\toprule
 & STDE & StyleFool & LocalStyleFool & Clean \\
\midrule
Naturalness & 2.32 & 3.06 & 3.45 & 3.60 \\
Consistency & 1.91 & 3.24 & 3.65 & 3.92 \\
\bottomrule
\end{tabular}
}
\vspace{-2mm}
\end{table}

\begin{table}[t]  
\centering
\caption{Results of realness test.}
% \vspace{-2mm}
\label{tab:realness}
\resizebox{0.9\linewidth}{!}{
\begin{tabular}{ccccc}
\toprule
 & LocalStyleFool is real & Another one is real & Both real & Both unreal \\
\midrule
LocalStyleFool-STDE & 51.5\% & 6.2\% & 10.3\% & 32.0\% \\
LocalStyleFool-StyleFool & 40.2\% & 20.6\% & 29.9\% & 9.3\% \\
LocalStyleFool-Clean & 34.0\% & 37.1\% & 26.8\% & 2.1\% \\
\bottomrule
\end{tabular}
}
\vspace{-2mm}
\end{table}

\subsection{Ablation Study}
LocalStyleFool considers transfer-based gradient information and regional area at the region selection stage. To explore the contribution of these two criteria to the attack, we conduct an ablation study. As a control, we randomly select regions in the original video before conducting style transfer. Then, we ablate the gradient information ($\alpha = 1$) and regional area restriction ($\alpha = 0$) respectively. We randomly select 50 videos from UCF-101 and attack the C3D model. Figure~\ref{fig:visualization_ablation} shows the visualization of an example. When the region is randomly selected or only based on the gradient information, the color and texture tend to be unnatural, which can cause alarm. Also, the attack becomes difficult or even vain if the regions selected based on gradient information are rather small. When only the regional area is considered, the selected areas may not overlap with the dominant regions of the original video, which requires an additional average of over 12\% queries. Overall, only when both two criteria are considered can LocalStyleFool achieve both high attack efficiency and sensory comfort, \eg the playing field seems to have only been renovated as shown in the last row. 

\section{Conclusion}
To address the problem of low naturalness in local areas in the existing style-transfer-based attack, we propose LocalStyleFool, which adds regional style-transfer-based perturbations to ameliorate the video quality based on SAM. We design an associative criterion of the combination of transfer-based gradient information and regional area to select the regions for style transfer and track these regions through the video stream. According to the user study, the adversarial videos generated by LocalStyleFool improve the imperceptibility in terms of naturalness, realness, and consistency while maintaining competent attack efficiency. SAM also helps LocalStyleFool to consume fewer queries in high-resolution data and avoid regional artifacts. Our work also exposes the negative aspect of SAM if misused for malicious purposes. We will plumb possible defenses for style-transfer-based attacks in the future.

\section{Acknowledgments}
This work was supported in part by the Overseas Research Cooperation Fund of Tsinghua Shenzhen International Graduate School (HW2021013), Guangdong Basic and Applied Basic Research Foundation (2022A1515010417) and the Key Project of Shenzhen Municipality (JSGG20211029095545002).

{
    \small
    \bibliographystyle{unsrt}
    \bibliography{ref}
}

\appendix
\subsection{Details for Experimental Setup}
In style transfer stage, we use the following loss to iteratively update the input video:
\begin{small}
\begin{equation}
\label{equation:total_loss}
\begin{split}
{L_{st}} = \sum {\left( {\alpha {L_{cont}} + \beta {L_{style}} + \gamma {L_{tv}} + \delta {L_{real}}} \right)} \\
+ \lambda \sum {{L_{temp}}},
\end{split}
\end{equation}
\end{small}%
where $L_{cont}$, $L_{style}$, $L_{tv}$, $L_{real}$ and $L_{temp}$ represents the content loss, style loss, total variance loss, realistic loss~\cite{luan2017deep} and temporal loss respectively, $\alpha $, $\beta$, $\gamma$, $\delta$ and $\lambda$ are weight coefficients. In our experiments, we first fine-tune the parameters to achieve improved aesthetic results for visualization. We finally adopt the weight parameters as follows: $\alpha = 100$, $\beta = 10^6$, $\gamma = 10^{-4}$, $\delta = 5$, $\lambda = 20$. These parameter values were carefully chosen to strike a balance between style fidelity, content preservation, smoothness, and temporal consistency in the resulting video, but they can also be slightly changed if the attackers wish. We use PWC-Net~\cite{sun2018pwc} in optical flow estimation. We choose layer 4 of the pre-trained ResNet-50 model~\cite{he2016ResNet} as the output layer. In frame segmentation, SAM~\cite{kirillov2023SAM} provides three models (vit-h, vit-l and vit-b), which differ in backbone sizes. We use the default vit-h model to ensure the best segmentation accuracy.

\subsection{Discussion}
\noindent\textbf{Other Video Attacks.}
Since LocalStyleFool is an unrestricted attack, it is unfair to compare it with a line of restricted attacks~\cite{jiang2019black,wang2021reinforcement}. We also do not compare our attack with universal attacks, which is another attack branch where the perturbations are first trained offline, and then applied to videos in the test/new data. Classic video universal attacks include C-DUP (white-box)~\cite{li2019stealthy} and U3D (black-box)~\cite{xie2022universal}. Universal attacks usually require high computational cost, since the attackers hope that adding trained universal perturbations to any video can cause misclassification. In most cases in this scenario, only untargeted attacks can be achieved. Therefore, due to different attack goals, it is also inappropriate to make comparisons between our attack and universal attacks. 

\noindent\textbf{Potential Mitigation.}
LocalStyleFool further unleashes the power of unrestricted attack by selecting local regions using SAM for perturbation superposition. In addition to the attack performance and naturalness of LocalStyleFool, its robustness against potential mitigations is also of interest to the attacker. Compared to attacks with $\ell_p$-bounded perturbations, proactive defenses based on adversarial training~\cite{madry2017towards}, or more general distributional robust optimization~\cite{kang2022certifying}, may offer little help in alleviating the local style-transfer-based attack. The reason is that it is a non-trivial task defining a proper adversarial risk to minimize, given the fact that the position and the style reference of the perturbations are unknown to the defender. On the other hand, considering that LocalStyleFool is superior to StyleFool in terms of consistency, consistency detection defenses (\eg AdvIT~\cite{xiao2019advit}) are even more unable to detect adversarial samples generated by LocalStyleFool. Image reconstruction defenses (\eg ComDefend~\cite{jia2019comdefend}) may demonstrate certain defensive performance against the attack. These methods generally work based on the intuition that image reconstruction techniques have a chance to erase the adversarial perturbation or project it back to the manifold of the clean data distribution. Nevertheless, compared to global perturbations, local perturbations could be more resistant towards these defenses since the small perturbed regions have a higher chance of surviving from mitigation methods using random crops or random noise augmentation. Moreover, properly crafted locally stylized perturbations may better preserve the original distribution of the data, rendering image-reconstruction-based defenses less effective. Last but not least, LocalStyleFool also poses a threat against provable defenses, such as randomized smoothing~\cite{cohen2019certified}. The key challenge in applying randomized smoothing to defend against LocalStyleFool is that the perturbations, though only occupying some of the pixels in the inputs, can have a large $\ell_p$ magnitude which exceeds the certifiable radius of randomized smoothing. Since LocalStyleFool adds different style transfer to different regions, there might be insufficient smoothness in the splicing area between two adjacent regions. Therefore, a potential mitigation could be to detect the smoothness of color transitions. To conclude, it still requires efforts to propose effective defenses against LocalStyleFool and style-transfer-based attacks.

\subsection{Future Prospect}
In recent years, DNN-based models have been widely employed in various video-related tasks, including action recognition, video understanding, \etc. Currently, the emergence of Large Language Models (LLMs) has prompted us to pioneer an investigation into the deleterious effects of the Segment Anything Model when employed to attack DNN-based video recognition models. While conducting attacks on real-world commercial models proves to be financially burdensome due to the substantial computational resources required — approximately $10^3 \sim 10^4$ queries per video in digital attacks — we posit that the practical feasibility of such endeavors may be realized with the advent of query-efficient attack methodologies in the near future. Notably, the formidable capabilities of LLMs have been demonstrated in the image domain, wherein semantically meaningful adversarial perturbations have been generated by GPT-4 under the intentional guidance of the attacker~\cite{liu2023instruct2attack}. Hence, an additional prospective avenue of research involves leveraging LLMs to actively generate adversarial videos, given their superior capacity for video comprehension compared to DNN-based models.

\end{document}